\documentclass[twocolumn,conference,letterpaper]{IEEEtran}
\usepackage[left=0.75in,right=0.75in,top=1in,bottom=0.75in]{geometry}

\usepackage{cite}
\usepackage{amssymb,amsmath,latexsym,amsfonts,amsthm,mathtools}
\DeclareMathOperator*{\argmax}{arg\,max}

\usepackage{graphicx}
\usepackage{float,subcaption}
\usepackage{textcomp}
\usepackage{xcolor}
\definecolor{darkpastelpurple}{rgb}{0.59, 0.44, 0.84}
\usepackage{hyperref}
\hypersetup{colorlinks, breaklinks, citecolor=blue, linkcolor=blue, urlcolor=blue}
\usepackage[nameinlink]{cleveref}
\Crefformat{figure}{#2Fig.~#1#3}
\Crefmultiformat{figure}{Figs.~#2#1#3}{ and~#2#1#3}{, #2#1#3}{ and~#2#1#3}

\usepackage{tikz}
\usetikzlibrary{automata, shapes, arrows, calc, arrows.meta, fit, positioning}

\theoremstyle{plain}

\newtheorem*{problem*}{Problem}
\theoremstyle{remark}

\theoremstyle{definition}

\usepackage{mathrsfs}

\usepackage{newtxmath}
\usepackage{booktabs}
\usepackage{duckuments}
\usepackage{bm}

\newcommand{\St}{{\mathcal X}}
\newcommand{\Ac}{{\mathcal U}}
\newcommand{\Tm}{{\mathcal P}}
\newcommand{\R}{{r}}
\newcommand{\Expect}{{\mathbb E}}

\IEEEoverridecommandlockouts

\begin{document}
	\title{Adaptive Event-triggered Reinforcement Learning Control for Complex Nonlinear Systems}
	\author{Umer Siddique, Abhinav Sinha,~\IEEEmembership{Member,~IEEE}, and Yongcan Cao,~\IEEEmembership{Senior Member,~IEEE}
		\thanks{U. Siddique and Y. Cao are with the Unmanned Systems Lab, Department of Electrical and Computer Engineering, The University of Texas at San Antonio, San Antonio, TX 78249, USA. (e-mails: muhammadumer.siddique@my.utsa.edu, yongcan.cao@utsa.edu). A. Sinha is with the GALACxIS Lab, Department of Aerospace Engineering and Engineering Mechanics, University of Cincinnati, OH 45221, USA. (email: abhinav.sinha@uc.edu).}
	}

	\maketitle
	\thispagestyle{empty}
	
	\begin{abstract}
		In this paper, we propose an adaptive event-triggered reinforcement learning control for continuous-time nonlinear systems, subject to bounded uncertainties, characterized by complex interactions. Specifically, the proposed method is capable of jointly learning both the control policy and the communication policy, thereby reducing the number of parameters and computational overhead when learning them separately or only one of them. By augmenting the state space with accrued rewards that represent the performance over the entire trajectory, we show that accurate and efficient determination of triggering conditions is possible without the need for explicit learning triggering conditions, thereby leading to an adaptive non-stationary policy. Finally, we provide several numerical examples to demonstrate the effectiveness of the proposed approach.
	\end{abstract}
	
	\begin{IEEEkeywords}
		Data-driven control, reinforcement learning, event-triggered control, nonlinear systems.
	\end{IEEEkeywords}
	
	\section{Introduction}\label{sec:introduction}
 Event-triggered control constitutes a paradigmatic approach to control system implementation, wherein data exchange between the plant and its controller is precipitated by the satisfaction of a state- or output-dependent criterion. The underlying rationale is to facilitate communication between the plant and its controller solely when necessitated by the pursuit of desired control objectives, diverging from traditional time-triggered (periodic) strategies that schedule communication instants based on elapsed time rather than actual system requirements. The motivation behind event-triggered control lies in its applicability to resource-constrained scenarios, where the costs associated with communication, computation, and control input updates are non-negligible, and the optimal sampling rate is unknown or varies over time, such as in networked control systems and embedded systems. 
 
 While there exists a multitude of works addressing the design of event-triggered controllers (see, for example, \cite{10.1201/b19013} and references therein), a vast majority of them require knowledge of the system model to be controlled. In practice, it is hard to model general nonlinear systems with complex interactions (e.g., when modeling via first principles may not be tractable and or a desired level of accuracy in system identification cannot be reached due to noisy data), and model-free control strategies appear alluring in such cases. A few previous works have presented data-driven techniques to improve and augment event-triggered control and state estimation, e.g.,  \cite{BATTISTELLI201875,HUANG2017256,8619335,8027115}. In these studies, learning is leveraged to approximate intractable conditional probability densities that emerge in decentralized optimization problems or to derive tractable solutions to Hamilton-Jacobi-Bellman equations that yield optimal control policies. This is achieved through various methodologies, including model-free approaches such as Q-learning, which eschew explicit system modeling in favor of iterative learning, as well as neural network-based methods that harness the representational power of artificial neural networks to approximate complex value functions or control policies. 
 
 Currently, only a few works are available for data-driven event-triggered control, e.g., \cite{10042042,https://doi.org/10.1002/rnc.6024,8619335, 10147364,9838043} where discrete-time formulation has been presented. On the contrary, the formulation in \cite{9849174} is continuous-time. It is worth noting that the learning process, particularly for linear time-invariant systems, simplifies in \cite{10042042,https://doi.org/10.1002/rnc.6024,9838043} by ignoring the effect of disturbances in the offline data collection process. Conversely, the approach proposed in \cite{10147364} adopts a more realistic paradigm, wherein the controller and triggering policy are designed solely based on a single batch of noisy data collected from the system, thereby introducing an additional layer of complexity due to the presence of disturbances and measurement errors. On the other hand, disturbances are taken into account during both the learning phase and closed-loop operation in the work of \cite{9849174}, where a dynamic triggering strategy is introduced to guarantee the preservation of $\mathcal{L}_2$ stability.
 
 As a sequential decision-making strategy, deep reinforcement learning (RL) has been shown to be successfully applied to many control problems in, e.g., robotics \cite{ibarz2021train}. However, the typical focus in such problems is on the design of control policies only while the consideration of communication cost is often overlooked. There are a few studies where model-free RL-based design has been presented for event-triggered control. For example, in \cite{VAMVOUDAKIS2018412}, an actor-critic method has been presented to learn an event-triggered controller with a predefined communication trigger. This essentially implies that the decision about when to communicate is not learned from scratch. The work in \cite{6889787} considered a fixed error threshold for communication triggering and leverages approximate dynamic programming to learn an event-triggered controller, whereas both the model of the system and an optimal event-triggered controller with fixed communication trigger have been simultaneously learned in \cite{8183439}. The work in \cite{8619335} proposed to learn both the communication and the control policies using Deep Deterministic Policy Gradients (DDPG) within the RL framework, which is only applicable to off-policy deep RL methods.

 Despite recent advances in data-driven control, there still arise several limitations, such as the presence of triggering conditions that often depend on simplified models or assumptions (and, therefore, may not accurately reflect real-world complexities), fixed conditions struggle to adapt in complex dynamic systems, and the reactive conditions (i.e., those depending on the current and previous state differences) that often lead to suboptimal behavior. Hence, there is a pressing need for the development of general, data-driven event-triggered control methodologies that can be seamlessly applied to continuous-time nonlinear systems in the presence of noise/perturbation, which are ubiquitous in real-world processes. These techniques must demonstrate robustness to disturbances that can affect both the data acquisition process and the closed-loop operations toward reliable performance in the face of uncertainty and noise. Furthermore, the separation principle does not generally hold in the case of event-triggered controllers. Meanwhile, the lack of coordination between triggering conditions and control policies, when trained separately, may result in inefficient or even conflicting actions. Hence, it is desired to learn both control and communication policies simultaneously.

 In this paper, we propose an adaptive event-triggered proximal policy optimization (ATPPO) method to jointly learn the triggering condition and the control policy in a continuous-time nonlinear system subject to bounded uncertainties. Unlike previous approaches, e.g., \cite{6889787,lu2023event,chen2022reinforcement}, that rely on manually designed or state-difference-based triggering conditions, our method integrates the learning of both aspects, hence reducing the number of parameters and computational overhead associated with separate learning processes. Moreover, the proposed method adaptively learns a non-stationary policy by considering accrued rewards and past information, yielding an optimized behavior over entire trajectories rather than just responding to the most recent state. By augmenting the state space with accrued rewards,  we facilitate the learning of value functions that encapsulate the agent's history, allowing for a more accurate and efficient determination of triggering conditions without the need for explicit training. Via several illustrative examples, we show that this joint optimization strategy represents a unique yet efficient way of learning event-triggered controllers. To the best of the authors' knowledge, it is the first method to employ a non-stationary policy for this purpose, leading to more cohesive and effective learning of communication and control policies in complex nonlinear systems.

\section{Background and Preliminaries} \label{sec:preliminaries}
We consider a general nonlinear system subject to bounded uncertainty,
\begin{equation}\label{sys}
    \dot{x} = f(x) + b(x)u + d,
\end{equation}
where $x\in \mathbb{R}^n$ is the state, $u\in \mathbb{R}^m$ is the control input at time $t\in\mathbb{R}_{\geq 0}$ and $d\in\mathbb{R}^n$ denotes the exogenous disturbance such that $\|d\|\leq d_{\max}$ for some $d_{\max}\in\mathbb{R}_+$. Without loss of generality, we assume that the nonlinear unknown function, $f(x)$, is locally Lipschitz, that is, $\|f(x_1)-f(x_2)\|\leq L\|x_1-x_2\|$ for some $x_1,x_2\in\mathbb{R}^n$, $L\in\mathbb{R}_+$, and for all time $t\in\mathbb{R}_{\geq 0}$. The goal is to learn a suitable policy $u=K(x)$ to stabilize \eqref{sys} while considering resource-aware scheduling of control and communication. The resource-aware controller is implemented using zero-order hold devices, leading to the control input $u(t) = u(t_k)$, for all $t\in[t_k,t_{k+1})$ such that the inter-event time $\Delta = t_{k+1}-t_k$ is not necessarily fixed for all $k\in\mathbb{N}\cup \{0\}$. Later, we show that $\Delta$ is lower bound by a positive constant, thus eliminating the possibility of the Zeno phenomenon.

Without loss of generality, we assume that the scheduling is initiated at $t=0$, thus $t_0=0$ and the sequence of event times can be expressed iteratively as $t_{k+1}=\inf\left\{t:~t>t_k,\Phi\left(t,x(t),\hat{x}(t)\right)\geq 0\right\}$ for some triggering condition $\Phi$, which is a function of the true state at the triggering instant $x(t)$ and the last broadcast state $\hat{x}(t)$. Then, one may note that in event-triggered scheduling, the control effort at the next triggering instant takes the form,
\begin{equation}
u(t) =
\begin{cases}
     u(t_{k}); & \text{if}~\Phi\left(t,x(t),\hat{x}(t)\right)\geq 0~\forall~t\in[t_k,t_{k+1}), \\
    u(t_{k-1}); & \text{otherwise},
\end{cases}
\end{equation}
where $u(t_{k}) = K(x(t_{k}))$ is the control computed at the triggering instant $t_k$,  $\Phi\geq 0$ is the triggering condition that decides when to schedule a control input, and $\hat{x}(t_k)$ is the last broadcast state. Specifically, one has $\hat{x}(t) = x(t_k),~\forall~t\in[t_k,t_{k+1})$. In the proposed approach, we aim to jointly learn both the control policy $u(t)$ as well as the triggering condition $\Phi$ (which essentially translates to the communication policy or \textit{when to communicate}). To this end, we present a brief overview of reinforcement learning below.

\subsection{Reinforcement Learning}
Reinforcement Learning (RL) is a framework where an autonomous agent learns to make decisions through interactions with an environment that is typically unknown. These interactions occur sequentially at each time step $t_k$, where the agent observes the current state $x(t_k)$, selects an action $u(t_k)$, receives a reward $r(t_k)$, and transitions to the next state $x(t_{k+1})$. This process is usually modeled as a Markov Decision Process (MDP) \cite{puterman1990markov}, which is defined by the tuple $\St, \Ac, \Tm, \R, \gamma$. Here, $\St$ and $\Ac$ are the state and action spaces, $\Tm$ is the state transition probability function, $\R$ is the reward function, and $\gamma \in [0,1) $ is the discount factor that determines the importance of future rewards.

The agent’s goal is to learn an optimal policy $ \pi $ that maximizes the expected cumulative discounted reward, $\Expect_{\pi} \left[\sum_{t=0}^\infty \gamma^{t} \R_{t} \right]$.
Here the expectation is taken over the stochastic outcomes of states and actions under the policy $ \pi $, which maps states to a probability distribution over actions. 
A policy can be either deterministic, $\pi(x) = u $, or stochastic, $ \pi(u|x) = \Tm_{\pi}[\Ac = u | \St = x] $. 
A policy is considered \textit{Markov} if it depends only on the current state, and \textit{stationary} if it remains consistent over time. 
Conversely, a \textit{non-stationary} policy allows for adaptation over time, potentially improving performance in non-stationary environments. 
This paper considers non-stationary and stochastic policies to address dynamic, nonlinear, and evolving environments.

To evaluate the quality of a given policy $\pi$, we define the state value function $ V^\pi(x) $, which represents the expected return starting from state $x$ under policy $\pi$, as $V^{\pi} (x) = \Expect_{\pi}\left[\sum_{t=0}^{\infty} \gamma^{t} r_{t} \mid x_0 = x\right]$. Similarly, the state-action value function, $ Q^\pi(x, u) $, represents the expected return starting from state $x$, taking action $u$, and following policy $\pi$, defined as $
Q^\pi (x,u) = \Expect_{\pi} \left[\sum_{t=0}^{\infty} \gamma^{t} r_{t} \mid x_0 = x, u_0 = u \right]$. In RL, the optimal policy $ \pi^\star $ is defined as the policy that maximizes the expected value across all states, $\pi^\star = \argmax_\pi V^\pi (x)$.

However, finding the optimal policy directly often requires complete knowledge of the environment's dynamics, which is typically unavailable in real-world applications. As a result, \emph{model-free} RL focuses on learning policies through trial-and-error interactions with the environment. Since, in high-dimensional and continuous state-action space problems, the state and or action spaces can be very large, function approximators such as deep neural networks (DNNs) are often employed. In the context of deep RL, those deep neural networks are used to approximate value functions and policies. Popular methods such as Deep Q-Network (DQN) \cite{mnih2015human} approximate the state-action value function using DNNs, while actor-critic methods like Proximal Policy Optimization (PPO) \cite{schulman2017proximal} optimize the policy directly for continuous or discrete action spaces tasks.

\section{Proposed Approach}
In this section, we introduce the adaptive event-triggered proximal policy optimization (ATPPO) method, which extends the standard proximal policy optimization (PPO) \cite{schulman2017proximal} algorithm to jointly learn both the control policy and the triggering condition. This unified approach addresses the limitations of traditional event-triggered control methods that rely on triggering conditions designed manually or based on state differences by integrating the learning of these aspects into a single framework. While our approach is general and can be applied to any RL method, we specifically choose PPO due to its robustness, stability, and strong empirical performance across various tasks \cite{huang202237}.

In traditional event-triggered control, the triggering condition is often manually designed or based on state differences and is treated separately from the control policy. 
ATPPO jointly learns both the control action $u(t)$ and the triggering condition $\Phi$ as part of the policy $\pi(\mathfrak{x}) $ where $\mathfrak{x}$ is the augmented state that includes accumulated rewards and past actions. By incorporating the triggering condition $\Phi$ within the policy, ATPPO optimizes the control strategy while dynamically determining when communication should occur, effectively reducing the number of parameters and computational overhead when learning them separately.

Similar to the PPO, ATPPO is a policy gradient method designed to directly optimize policies in deep RL.
In other words, it directly explores the policy space instead of learning a Q-function, which is typically defined by parameterized functions, such as deep neural networks. 
The parametric policy is denoted as $\pi_\theta$, where $\theta$ represents the policy parameters. 
Thus, the goal of ATPPO is to maximize the expected sum of rewards akin to the standard RL goal, however with an additional penalty mechanism to discourage excessive triggering. 
If the triggering condition $\Phi$ is always true, meaning the agent tends to communicate at every time step, a penalty is imposed to encourage the agent to make more judicious decisions about when to trigger communication. 
This penalty promotes more efficient use of communication resources by optimizing the balance between control performance and communication frequency.

Formally, in ATPPO, the policy is trained to learn both the control action and the triggering condition, that is,
\begin{align*}
    (\Phi, u) = \pi(\mathfrak{x}),
\end{align*}
where $\mathfrak{x} = (x, r_{\text{acc}})$ represents the augmented state. The augmentation of the state with accrued rewards allows the agent to consider historical information when making decisions, enabling a more comprehensive and accurate determination of both control actions and triggering conditions. Thus, the optimization objective in ATPPO can be expressed as
\begin{align}
J(\theta) =& \Expect_{u \sim \pi_{\theta}(\cdot \mid \mathfrak{x})} \left[ \min(\rho_{\theta}  A(\mathfrak{x},u), \bar{\rho}_{\theta}  A(\mathfrak{x},u)) \right. \nonumber\\
&\left.- \Psi \cdot \mathbb{I}(\Phi \geq 0) \right],
\end{align}
where $\bar{\rho}_{\theta} = \text{clip}(\rho_{\theta}, 1 - \epsilon, 1 + \epsilon)$, $\rho_{\theta} = \dfrac{\pi_{\theta}(u|\mathfrak{x})}{\pi_b(u|\mathfrak{x})}$, $\pi_b$ represents the policy generating the transitions, $\epsilon$ is a hyperparameter that controls the constraint, and $\Psi$ is a hyperparameter controlling the penalty for frequent triggering. The indicator function $\mathbb{I}(\Phi \geq 0)$ applies the penalty when the triggering condition is met. 
Here, $A$ denotes an advantage function that quantifies the relative value of taking a specific action $u$ in the current state $\mathfrak{x}$ under the policy $\pi_{\theta}$. 

Since we learn with a gradient-based approach, learning or estimating the Q-value function can have a higher variance as it estimates the expected total reward from taking a specific action in a given state and then following the current policy. To reduce the variance, the advantage function is used, which subtracts the state value $V(\mathfrak{x}_t)$ and helps cancel out some of the variability that is common to all actions in a given state. In ATPPO, the advantage $A$ is estimated using $\lambda$-returns \cite{SchulmanWolskiDhariwalRadfordKlimov17}, and can be defined as
\begin{align}
    A(\mathfrak{x},u) = \sum_{t=1} (\gamma \lambda)^{t-1} r_t + \gamma V_{\phi}(\mathfrak{\Tilde{x}}) - V_{\phi}(\mathfrak{x}) \,,
\end{align}
where $\mathfrak{\Tilde{x}}=\mathfrak{x}+dt$ is the next state. The advantage function and value function are learned with the augmented state that enhances the ability of the advantage function and value function to capture the agent’s history, allowing for more informed decisions that consider long-term consequences rather than just immediate rewards. This approach also leads to the learning of non-stationary policies that optimize behavior over entire trajectories, leading to more cohesive and effective communication and control strategies in complex, dynamic environments.

\section{Environments for Experiments}
To demonstrate the effectiveness of ATPPO, we conduct experiments in various environments that include perturbed single integrator dynamics,  MuJoCo environments \cite{todorov2012mujoco} (such as Half-Cheetah, Hopper, Reacher), and in a target capture scenario where a pursuer captures a moving target \cite{siddique2024deep}. Below, we first provide the details of each environment, followed by the results and discussion in the next section.
\subsection{Single Integrator}  
In our initial experimental phase, we conducted experiments on a single integrator dynamics perturbed by a bounded uncertainty, which serves as a preliminary benchmark due to its simple dynamics that models a wide class of phenomena. This environment can be represented as
\begin{align}\label{eq:singleenv}
    \dot{x} = u + d,
\end{align}
where $\|d\|\leq d_{\max}$. In this environment, the agent is tasked with stabilizing the system in the presence of uncertainties, that is, to drive the system's state to the desired equilibrium point. Specifically, the state variable is required to converge to the origin from an initial value, $x(0)=5$. In this situation, the agent receives the current state $s_t$ and takes an action $a_t$ to adjust the state incrementally over time, with the goal of minimizing the distance between the current state and the origin (the equilibrium state). The reward function encourages the agent to reduce the absolute value of the state (i.e., getting closer to the equilibrium point) while penalizing large or unnecessary control actions. This environment favors intuition, simplicity of presentation, and a framework for validating the agent's learning capacity. Specifically, it allows us to test the proposed ATPPO algorithm in a minimalistic setting, focusing on how efficiently the agent learns to stabilize the system using optimal event-triggered control actions. The transition dynamics are governed by~\eqref{eq:singleenv}, where the agent interacts with the system through control actions that influence the state evolution.

\subsection{Half-Cheetah}
The Half-Cheetah environment represents a planar biped robot with a 17-dimensional state space and a 6-dimensional action space. The state vector comprises joint angles and velocities, while the action space consists of joint torques. The reward function maximizes the forward velocity and minimizes control effort, which in other words, encourages the cheetah to move faster while penalizing control effort. The transition function governs the dynamics of a bipedal robot to move to the next state and provides a realistic scenario for evaluating locomotion efficiency.

\subsection{Hopper}
The Hopper environment is highly nonlinear and models a monopedal robot with an 11-dimensional state space and a 3-dimensional action space. The state vector includes body positions and velocities, and the action space consists of joint torques. The reward function is based on forward velocity and control costs. The transition function includes the dynamics of the monopedal about how it moves, allowing for realistic simulations of hopping and balance.

\subsection{Reacher} 
The Reacher environment simulates a two-degree-of-freedom robotic arm with nonlinear dynamics. Its state space includes arm joint angles, velocities, and target position, while the action space is a 2-dimensional vector of joint torques. The reward function is based on the distance between the end effector and the target. In this environment, the state transitions are governed by physics-based simulations, which model the complex nonlinear dynamics of the robotic arm which ensures realistic control and target reaching tasks.

\subsection{Target Capture}
In the target capture case, we consider one or more pursuers (denoted as P$_i$) and one or more targets (denoted as T$_j$). Each vehicle $k$ ($\forall k=\{\mathrm{P}_i,\mathrm{T}_j\} \mid i,j \in \mathbb{N}$) is nonholonomic, with its motion described by the following equations:
\begin{align}
    \dot{X}_k &= v_k \cos\psi_k, \\
    \dot{Y}_k &= v_k \sin\psi_k, \\
    \dot{\psi}_k &= \frac{a_k}{v_k},
\end{align}

\begin{figure*}[t]
\centering
    \begin{subfigure}[t]{0.32\textwidth}
    \centering
    \includegraphics[width=\linewidth]{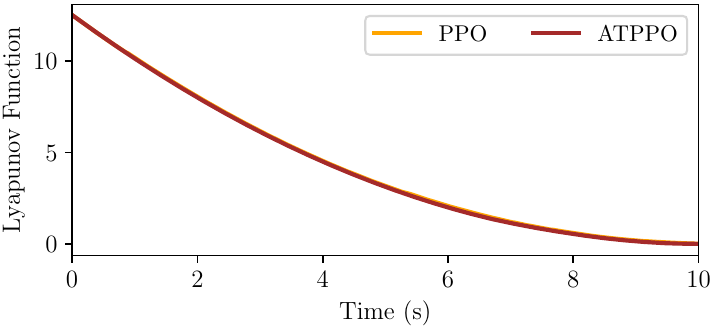}
    \caption{Lyapunov function.}
    \label{fig:sintlyap}
    \end{subfigure}
    \begin{subfigure}[t]{0.32\textwidth}
    \centering
    \includegraphics[width=\linewidth]{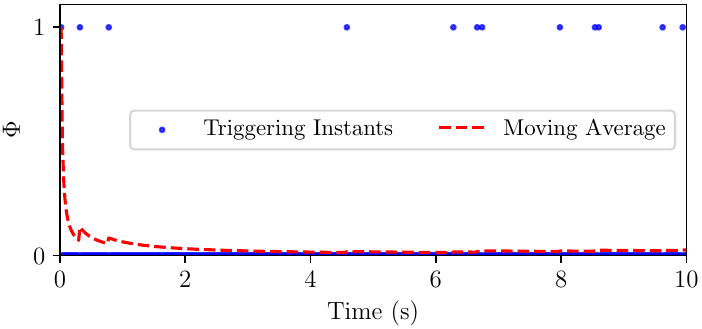}
    \caption{Triggering instants and moving avg.}
    \label{fig:sintcomm}
    \end{subfigure}
    \begin{subfigure}[t]{0.32\textwidth}
    \centering
    \includegraphics[width=\linewidth]{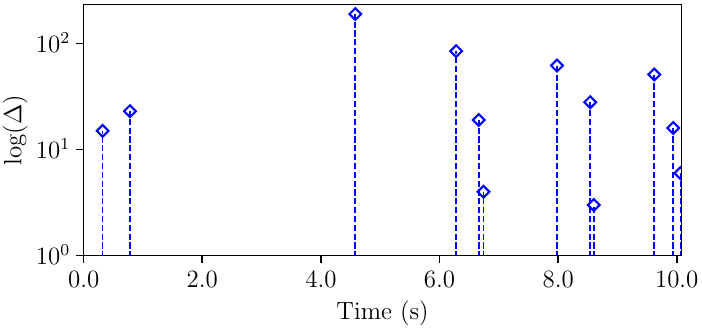}
    \caption{Inter event time.}
    \label{fig:sintzeno}
    \end{subfigure}
\caption{Performance comparison between PPO and ATPPO for a perturbed single integrator.} 
\label{fig:sint}
\end{figure*}

where $[X_k, Y_k]^\top \in \mathbb{R}^2$ represents the instantaneous position, $\psi_k \in (-\pi, \pi]$ is the heading angle, and $v_k$ is the speed of the $k$\textsuperscript{th} vehicle. The lateral acceleration $a_k$ steers the vehicle and accounts for its turning constraints. The system includes a first-order autopilot lag model to simulate real-world conditions:
\begin{equation}
    \frac{a_{\mathrm{P}_i}~\text{achieved}}{a_{\mathrm{P}_i}~\text{commanded}} = \frac{1}{1 + s\tau_i},
\end{equation}
where $\tau_i$ is the corresponding time constant.

For pursuer-target pairs, it is often beneficial to express these dynamics in a relative frame of reference, particularly in scenarios where absolute measurements are challenging or costly to obtain. The kinematics of relative motion between a pursuer $P_i$ and a target $T_j$ can be described as
\begin{subequations}\label{eq:engdyn}
		\begin{align}
			v_{r_{ij}} = \dot{r}_{ij} &= v_{\mathrm{T}_j} \cos\left(\psi_{\mathrm{T}_j} - \eta_{ij}\right) - v_{\mathrm{P}_i} \cos\sigma_{\mathrm{P}_{ij}}, \\
			v_{\eta_{ij}}  = r_{ij}\dot{\theta}_{ij} &= v_{\mathrm{T}_j} \sin\left(\psi_{\mathrm{T}_j} - \eta_{ij}\right) - v_{\mathrm{P}_i} \sin\sigma_{\mathrm{P}_{ij}}, \\
			\sigma_{\mathrm{P}_{ij}} &= \psi_{\mathrm{P}_i} - \eta_{ij},
		\end{align}
	\end{subequations}
where $v_{r_{ij}}$ and $v_{\eta_{ij}}$ are the components of the relative velocities along and across the line-of-sight, respectively. The relative distance $r_{ij}$ and the line-of-sight angle $\eta_{ij}$ between the pursuer and target are given by:
\begin{align}
    r_{ij} &= \sqrt{\left(X_{\mathrm{T}_j} - X_{\mathrm{P}_i}\right)^2 + \left(Y_{\mathrm{T}_j} - Y_{\mathrm{P}_i}\right)^2}, \\
    \eta_{ij} &= \tan^{-1}\frac{Y_{\mathrm{T}_j} - Y_{\mathrm{P}_i}}{X_{\mathrm{T}_j} - X_{\mathrm{P}_i}}.
\end{align}

We assume constant vehicle speeds $v_{\mathrm{P}_i}$ and $v_{\mathrm{T}_j}$, with $v_{\mathrm{T}_j} = \alpha v_{\mathrm{P}_i}$ and $\alpha \in [0, 1)$. Lateral accelerations $a_{\mathrm{P}_i}$ and $a_{\mathrm{T}_j}$ are limited by maximum bounds, $a_{\mathrm{P}_{\max}}$ and $a_{\mathrm{T}_{\max}}$, respectively.

To formulate the target capture problem as an MDP, we define the pursuer $P_i$ as an RL agent, the environment as the dynamics described above, the state vector as $\St_i = [r_{ij}, \dot{r}_{ij}, \eta_{ij}, \dot{\eta}_{ij}, \psi_{\mathrm{P}_i}]^\top$, and the action as the lateral acceleration $\Ac_i \in [a_{\mathrm{P}_{\max}}, a_{\mathrm{P}_{\min}}]$.  
The rewards function in this environment is defined as a convex combination of multifaceted objectives such that
\begin{equation}
    \R = \alpha_1 r_1+ \alpha_2 r_2 + \alpha_3 r_3 + \alpha_4 r_4 + \alpha_5 r_5,
\end{equation}
where $\displaystyle\sum_k \alpha_k=1$, and individual rewards are  
\begin{align}
    r_1 =& -\dfrac{r}{r(0)}, \\
    r_2 =& -\left(\dfrac{a_\mathrm{P}}{a_\mathrm{P}^{\max}}\right)^2, \\
    r_3 =& \begin{cases}
        \phantom{-}0.25 & \text{if}~v_r<0,\\
        \phantom{-}1 & \text{if}~v_{r}<0~\text{and}~v_{\eta}=0,\\
        -1 & \text{otherwise},
    \end{cases}\\
    r_4 =&  \begin{cases}
        100 & \text{if}~{r}<r_\mathrm{miss},\\
        0 & \text{otherwise},
    \end{cases}\\
    r_5 =&  \begin{cases}
        \max\left\{0,~ m\left(\dfrac{r_\mathrm{miss}-r}{r_\mathrm{miss}}\right)\right\} & \text{if}~{r}<r_\mathrm{miss},\\
        0 & \text{otherwise},
    \end{cases}
\end{align}
where $m>0$ is a tuning parameter and $r_\mathrm{miss}$ is the capture radius. The proposed reward function enables us to shape the pursuer's behavior in nuanced ways, aligning its actions with our domain-specific requirements and constraints. By judiciously assigning rewards, we can induce desirable properties such as safety, or energy efficiency, imbuing the agent with a sense of responsibility and principled decision-making.

\begin{figure*}[t]
\centering
    \begin{subfigure}[t]{0.32\textwidth}
    \centering
    \includegraphics[width=\linewidth]{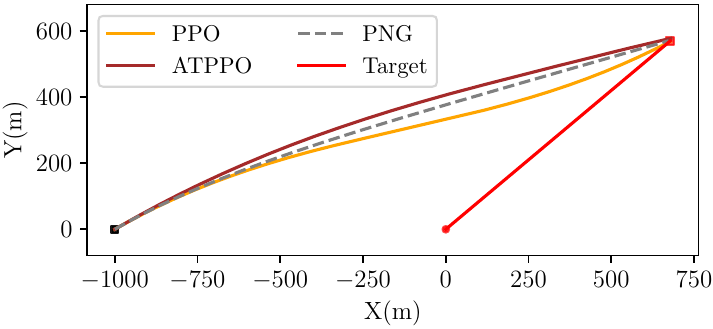}
    \caption{Trajectories.}
    \label{fig:engtraj}
    \end{subfigure}
    \begin{subfigure}[t]{0.32\textwidth}
    \centering
    \includegraphics[width=\linewidth]{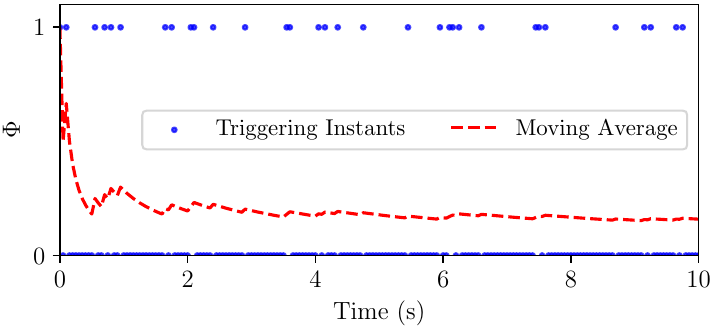}
    \caption{Triggering instants and moving avg.}
    \label{fig:engcomm}
    \end{subfigure}
    \begin{subfigure}[t]{0.32\textwidth}
    \centering
    \includegraphics[width=\linewidth]{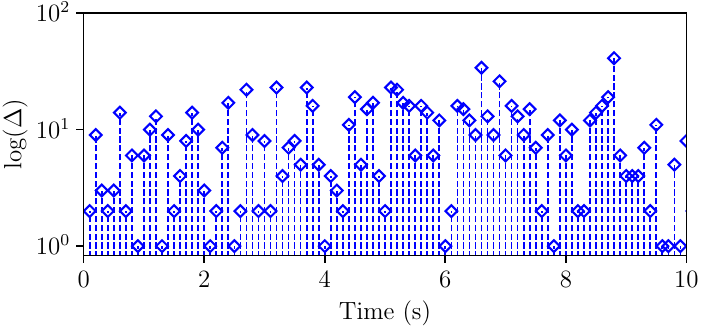}
    \caption{Inter event time.}
    \label{fig:engzeno}
    \end{subfigure}
\caption{Performance comparison between PPO and ATPPO in the pursuit-evasion environment.}
\label{fig:enggeo}
\end{figure*}

\section{Results}
In this section, we present the results of our experiments across various environments to demonstrate the effectiveness of the proposed ATPPO method. Through these experiments, we try to answer the following questions: \textbf{(A)} Why is event-triggered learning necessary, and how does it help to save resources while stabilizing complex nonlinear systems? \textbf{(B)} How does the proposed ATPPO method perform in complex, nonlinear, and highly uncertain robotic environments (such as those that require rendezvous or target capture)? \textbf{(C)} Can the proposed ATPPO approach generalize to high-dimensional, multi-degree-of-freedom MuJoCo tasks?

\begin{figure*}[t]
\centering
    \begin{subfigure}[t]{0.32\textwidth}
    \centering
    \includegraphics[width=\linewidth]{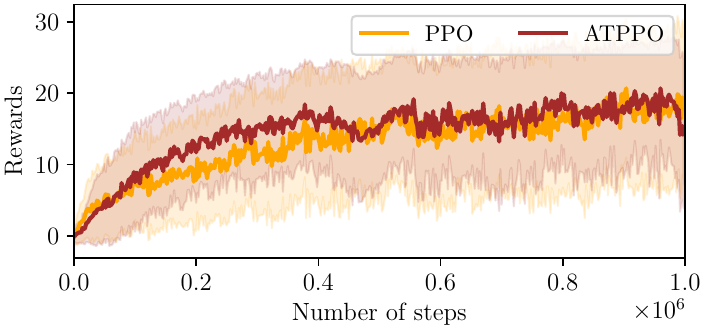}
    \caption{Rewards.}
    \label{fig:cheetahrewards}
    \end{subfigure}
    \begin{subfigure}[t]{0.32\textwidth}
    \centering
    \includegraphics[width=\linewidth]{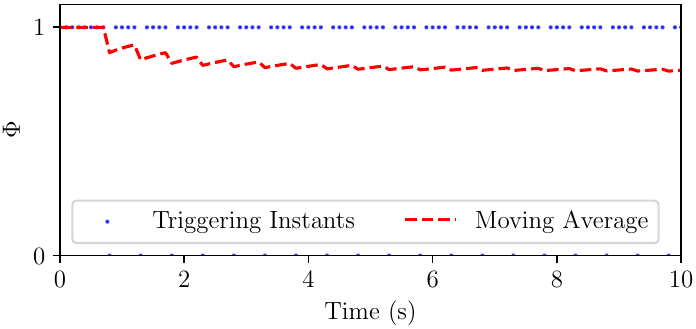}
    \caption{Triggering instants and moving avg.}
    \label{fig:cheetahcomm}
    \end{subfigure}
    \begin{subfigure}[t]{0.32\textwidth}
    \centering
    \includegraphics[width=\linewidth]{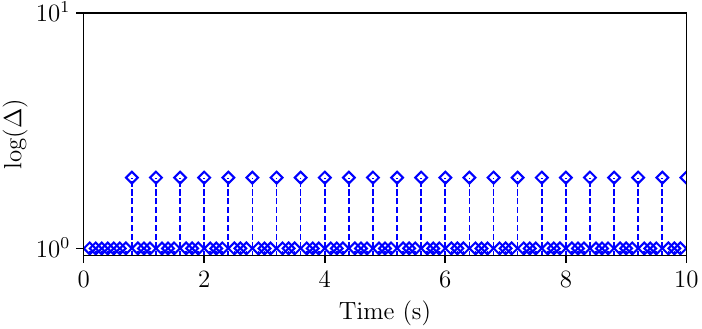}
    \caption{Inter event time.}
    \label{fig:cheetahzeno}
    \end{subfigure}
\caption{Performance comparison between PPO and ATPPO in the half-cheetah.}
\label{fig:cheetah}
\end{figure*}

\textbf{Question (A)} To answer this question, we first perform experiments on a perturbed single integrator. The goal of this experiment is to investigate how ATPPO's resource-aware scheduling mechanism minimizes the frequency of control actions, thereby saving communication and computation resources while still effectively stabilizing the system to the equilibrium state. \Cref{fig:sint} presents the performance comparison between PPO and the proposed ATPPO method for this case. The initial condition $x(0)=5$, and the agent is tasked to drive the initial state to the equilibrium $x(0)=0$ to stabilize the system. The results show that, compared to standard PPO, ATPPO is able to stabilize the system while saving at least $80$\% of resources and requires significantly less communication. \Cref{fig:sintlyap} shows the decay of the Lyapunov function $0.5x^2$ with over time. Both PPO and ATPPO allow for the gradual reaching of the equilibrium point, demonstrating their ability to stabilize the system. \Cref{fig:sintcomm} presents the communication frequency with the policy, showing that ATPPO significantly reduces the number of communication events compared to PPO. Despite fewer triggering instances, the system remains stable, indicating that ATPPO effectively balances resource savings and performance. Finally, \Cref{fig:sintzeno} shows the inter-event time between consecutive decisions and confirms that it remains non-zero and varies over time, ensuring that the triggering frequency is dynamically adjusted without leading to infinite triggerings.

\begin{figure*}[t]
\centering
    \begin{subfigure}[t]{0.32\textwidth}
    \centering
    \includegraphics[width=\linewidth]{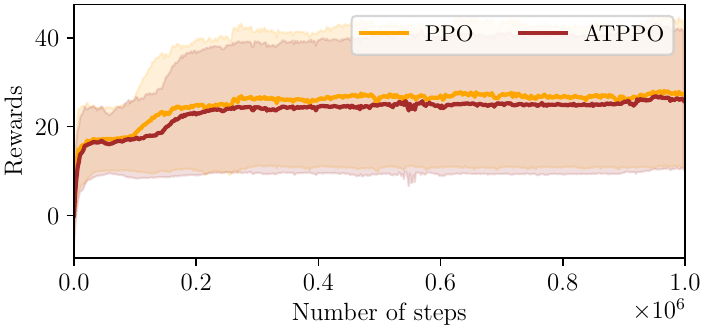}
    \caption{Rewards.}
    \label{fig:hopperrewards}
    \end{subfigure}
    \begin{subfigure}[t]{0.32\textwidth}
    \centering
    \includegraphics[width=\linewidth]{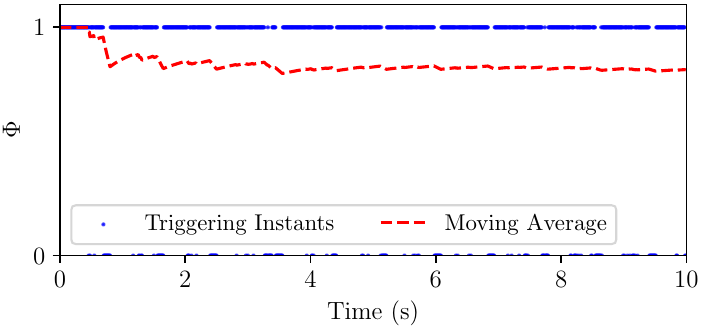}
    \caption{Triggering instants and moving avg.}
    \label{fig:hoppercomm}
    \end{subfigure}
    \begin{subfigure}[t]{0.32\textwidth}
    \centering
    \includegraphics[width=\linewidth]{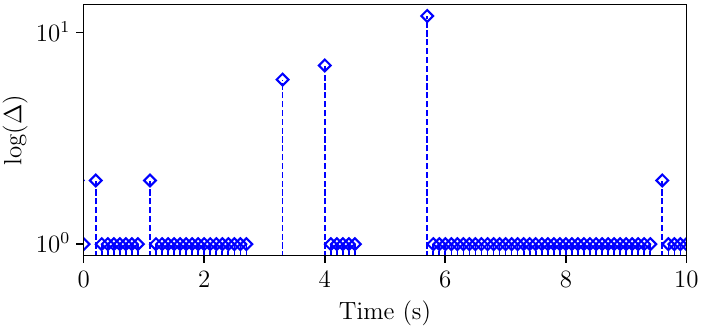}
    \caption{Inter event time.}
    \label{fig:hopperzeno}
    \end{subfigure}
\caption{Performance comparison between PPO and ATPPO in Hopper.}
\label{fig:hopper}
\end{figure*}
\begin{figure*}[t]
\centering
    \begin{subfigure}[t]{0.32\textwidth}
    \centering
    \includegraphics[width=\linewidth]{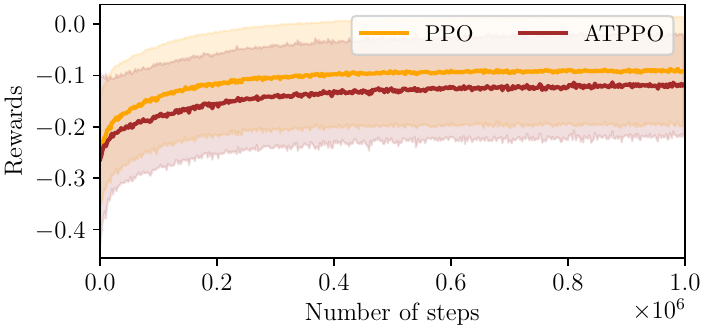}
    \caption{Rewards.}
    \label{fig:reacherrewards}
    \end{subfigure}
    \begin{subfigure}[t]{0.32\textwidth}
    \centering
    \includegraphics[width=\linewidth]{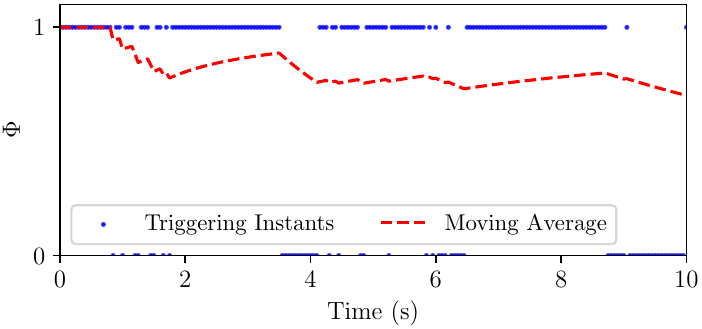}
    \caption{Triggering instants and moving avg.}
    \label{fig:reachercomm}
    \end{subfigure}
    \begin{subfigure}[t]{0.32\textwidth}
    \centering
    \includegraphics[width=\linewidth]{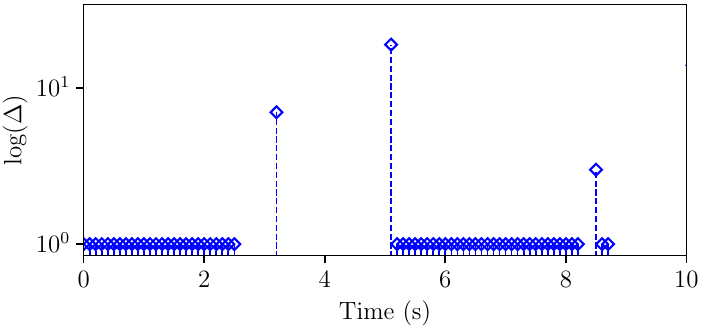}
    \caption{Inter event time.}
    \label{fig:reacherzeno}
    \end{subfigure}
\caption{Performance comparison between PPO and ATPPO in the Reacher.}
\label{fig:reacher}
\end{figure*}

\textbf{Question (B)} To answer this question, we conduct experiments where a pursuing vehicle needs to rendezvous with a target vehicle, which is moving. \Cref{fig:enggeo} shows the performance comparison of PPO and ATPPO in capturing a single target moving with a constant speed of 20 m/s and a heading angle of $40^\circ$. The pursuer has an initial speed of 40 m/s and starts 1000 m radially away from the target. The pursuer's lateral acceleration is limited to $\pm 5$ m/s$^2$, with an initial heading angle of $30^\circ$ and a line-of-sight angle of $0^\circ$. We also account for an autopilot lag with a time constant of 0.25 s. \Cref{fig:engtraj} shows that both agents successfully capture the target. However, the pursuer guided by ATPPO manages to arrive at the collision triangle by using fewer communications, and its performance closely resembles that of the proportional-navigation guidance (PNG), which is known for its optimality in intercepting non-maneuvering targets. In terms of communication frequency, \Cref{fig:engcomm} 
highlights ATPPO’s significantly lower communication frequency compared to PPO, yet it still ensures target capture.
These results demonstrate that ATPPO may be a suitable choice in a practical, complex, stochastic environment with minimal resource usage. Finally, \Cref{fig:engzeno} confirms the dynamic nature of ATPPO’s triggering mechanism, where no two communication decisions occur simultaneously, thereby avoiding Zeno behavior.

\textbf{Question (C)}  In order to answer this question, we conduct experiments in the MuJoCo environments \cite{todorov2012mujoco}, specifically focusing on the Half-Cheetah, Hopper, and Reacher environments. \Cref{fig:cheetah} presents the results of PPO and ATPPO in the Half-Cheetah environment. \Cref{fig:cheetahrewards} shows that both PPO and ATPPO achieve similar rewards across 1e6 timesteps. However, as shown in \Cref{fig:cheetahcomm}, ATPPO significantly reduces the number of communications. The blue dots represent discrete triggering events, while the red curve shows the moving average, which consistently decreases over the trajectory. This trend highlights the efficiency of ATPPO in reducing communication. \Cref{fig:cheetahzeno} verifies that the inter-event time between consecutive trigger decisions is always positive, confirming that ATPPO avoids Zeno behavior. For clarity, we display the inter-event time logarithmically. \Cref{fig:hopper} illustrates the performance of PPO and ATPPO in the Hopper environment, whereas \Cref{fig:hopperrewards} shows the rewards accumulated by both agents during training. While both methods improve steadily, achieving nearly identical rewards, ATPPO uses fewer communications, as shown in \Cref{fig:hoppercomm}.  Although ATPPO achieves slightly lower rewards, it demonstrates the advantage of reduced communication without significant performance loss. \Cref{fig:hopperzeno} further confirms the absence of Zeno behavior by showing consistently positive inter-event times throughout. Similarly, \Cref{fig:reacher} depicts the results in the Reacher environment. \Cref{fig:reacherrewards} demonstrates that both methods improve over time, although PPO ultimately achieves slightly higher rewards. However, as indicated in~\Cref{fig:reachercomm}, ATPPO significantly reduces communication frequency. \Cref{fig:reacherzeno} once again confirms that ATPPO maintains positive time intervals between communication events, ensuring no simultaneous triggers occur.

\section{Conclusions and Future Work}\label{sec:conclusions}
In this paper, we introduced the adaptive event-triggered proximal policy optimization (ATPPO) approach, a novel method for jointly learning control and communication policies in continuous-time non-linear systems subject to bounded uncertainties. By integrating the learning of both policies and augmenting the state space with accrued rewards, ATPPO enables the learning of a non-stationary policy that optimizes behavior over entire trajectories. This approach facilitates more accurate and efficient determination of triggering conditions without explicit training while reducing computational complexity and parameter count. Through several illustrative simulations, we demonstrated ATPPO's effectiveness in complex nonlinear systems where resource efficiency is crucial, showcasing its potential for real-world applications. Our work represents a unique and efficient approach to learning event-triggered controllers, offering a more cohesive method for simultaneously optimizing communication and control policies in complex systems. As a future work, we plan to extend ATPPO to multi-agent RL in both centralized and distributed settings. 
	
\bibliographystyle{IEEEtran}
\bibliography{references}

\end{document}